\documentclass[lettersize,journal]{IEEEtran}
\usepackage{amsmath,amsfonts}
\usepackage{algorithmic}
\usepackage{array}
\usepackage{subfigure}
\usepackage[caption=false,font=normalsize,labelfont=sf,textfont=sf]{subfig}
\usepackage{cite}
\usepackage{textcomp}
\usepackage{stfloats}
\usepackage{url}
\usepackage{microtype}
\usepackage{arydshln}
\usepackage[font=footnotesize]{caption}
\usepackage{verbatim}
\usepackage{graphicx}
\usepackage{tabularray,rotating,booktabs}
\usepackage{xspace}
\usepackage{xcolor}
\usepackage{balance}
\def\BibTeX{{\rm B\kern-.05em{\sc i\kern-.025em b}\kern-.08em
    T\kern-.1667em\lower.7ex\hbox{E}\kern-.125emX}}

\begin{document}
\title{Learning from Mistakes: Self-Regularizing Hierarchical Representations\\in Point Cloud Semantic Segmentation}
\author{Elena Camuffo,~\IEEEmembership{Student Member,~IEEE}, Umberto Michieli, Simone Milani,~\IEEEmembership{Member,~IEEE}
\thanks{All the authors are with the University of Padova.\protect\\
Elena Camuffo is the corresponding author.\protect\\
E-mail: \{elena.camuffo, umberto.michieli, simone.milani\}@dei.unipd.it}}


\newcommand{\revv}[1]{{\leavevmode\color{black} #1}}
\newcommand{\rev}[1]{{\leavevmode\color{black} #1}}
\newcommand{\del}[1]{{\leavevmode\color{red} #1}}

\newcommand{\um}[1]{{\leavevmode\color{orange}\textbf{UM:} #1}}
\newcommand{\sm}[1]{{\leavevmode\color{black} #1}} 
\newcommand{\smtwo}[1]{{\leavevmode\color{magenta} #1}} 

\makeatletter
\DeclareRobustCommand\onedot{\futurelet\@let@token\@onedot}
\def\@onedot{\ifx\@let@token.\else.\null\fi\xspace}
\def\eg{\emph{e.g}\onedot} \def\Eg{\emph{E.g}\onedot}
\def\ie{\emph{i.e}\onedot} \def\Ie{\emph{I.e}\onedot}
\def\cf{\emph{cf}\onedot} \def\Cf{\emph{Cf}\onedot}
\def\etc{\emph{etc}\onedot} \def\vs{\emph{vs}\onedot}
\def\wrt{w.r.t\onedot} \def\dof{d.o.f\onedot}
\def\iid{i.i.d\onedot} \def\wolog{w.l.o.g\onedot}
\def\etal{\emph{et al}\onedot}
\makeatother

\maketitle

\begin{abstract}
   Recent advances in autonomous robotic technologies have highlighted the growing need for precise environmental analysis. \rev{Point cloud} semantic segmentation has gained attention to accomplish fine-grained scene understanding by acting directly on raw content provided by sensors.
      {Recent solutions showed how different learning techniques can be used to improve the performance of the model, without any architectural or dataset change.}
      {Following this trend, we present a coarse-to-fine setup that LEArns from classification mistaKes (LEAK) derived from a standard model.}
   {First, classes are clustered into macro groups according to mutual prediction errors; then, the learning process is regularized by: (1) aligning class-conditional prototypical feature representation for both fine and coarse classes, (2) weighting instances with a per-class fairness index.}
  Our LEAK approach is very general and can be seamlessly applied on top of any segmentation architecture; indeed, experimental results showed that it enables state-of-the-art performances on different architectures, datasets and tasks, while ensuring more balanced class-wise results and faster convergence.
\end{abstract}

\begin{IEEEkeywords}
Point Clouds, Semantic Segmentation, Representation Learning, Spectral Clustering, Prototypes, Fairness.
\end{IEEEkeywords}

\section{Introduction}
\label{sec:intro}

\begin{figure}[t]
    \centering
    \includegraphics[trim=0.5cm 0.2cm 0cm 0.5cm, clip, width=0.9\linewidth]{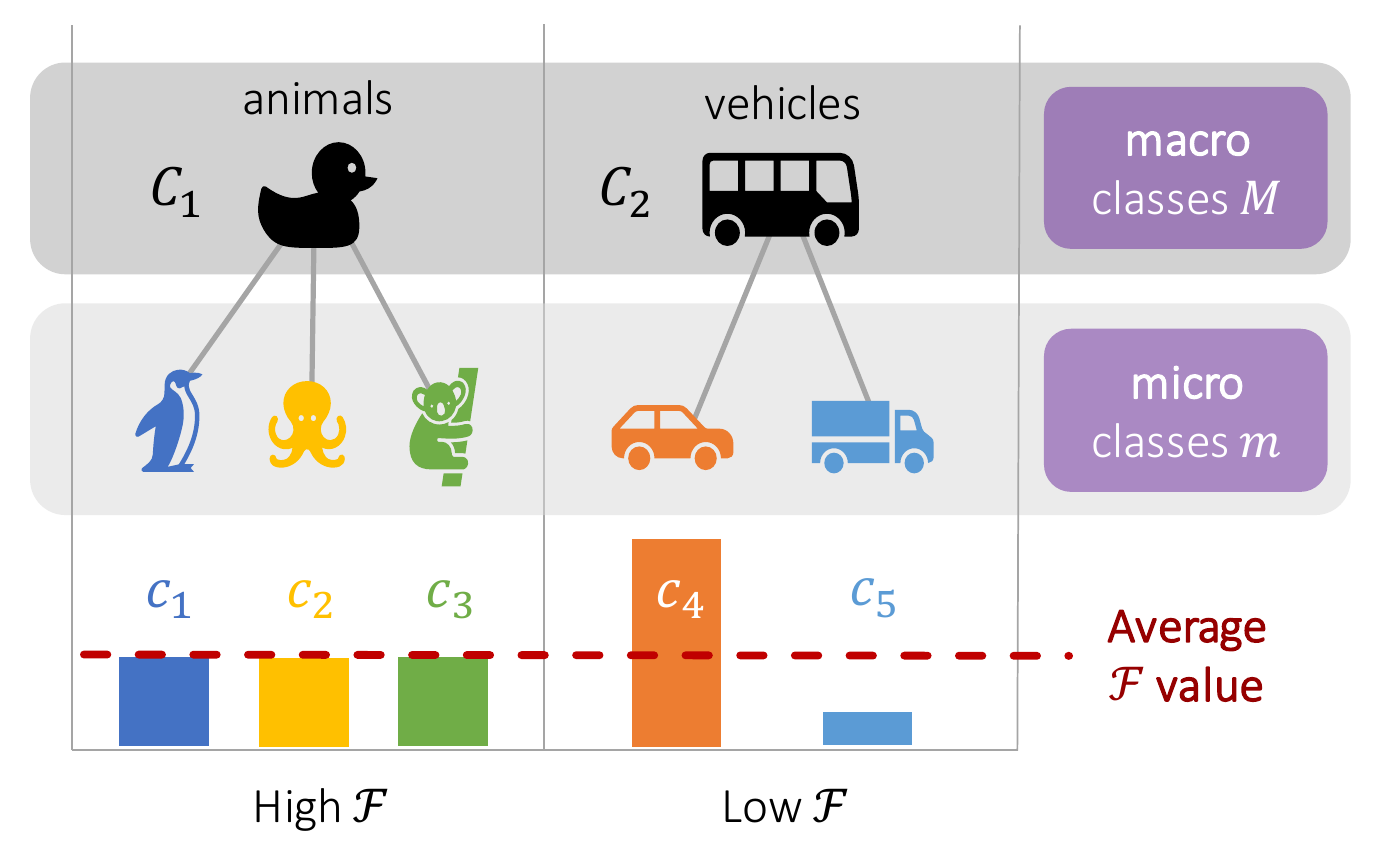}
    \vskip 2ex
     \includegraphics[trim=0.3cm 0.1cm 0cm 0.15cm, clip,width=\linewidth]{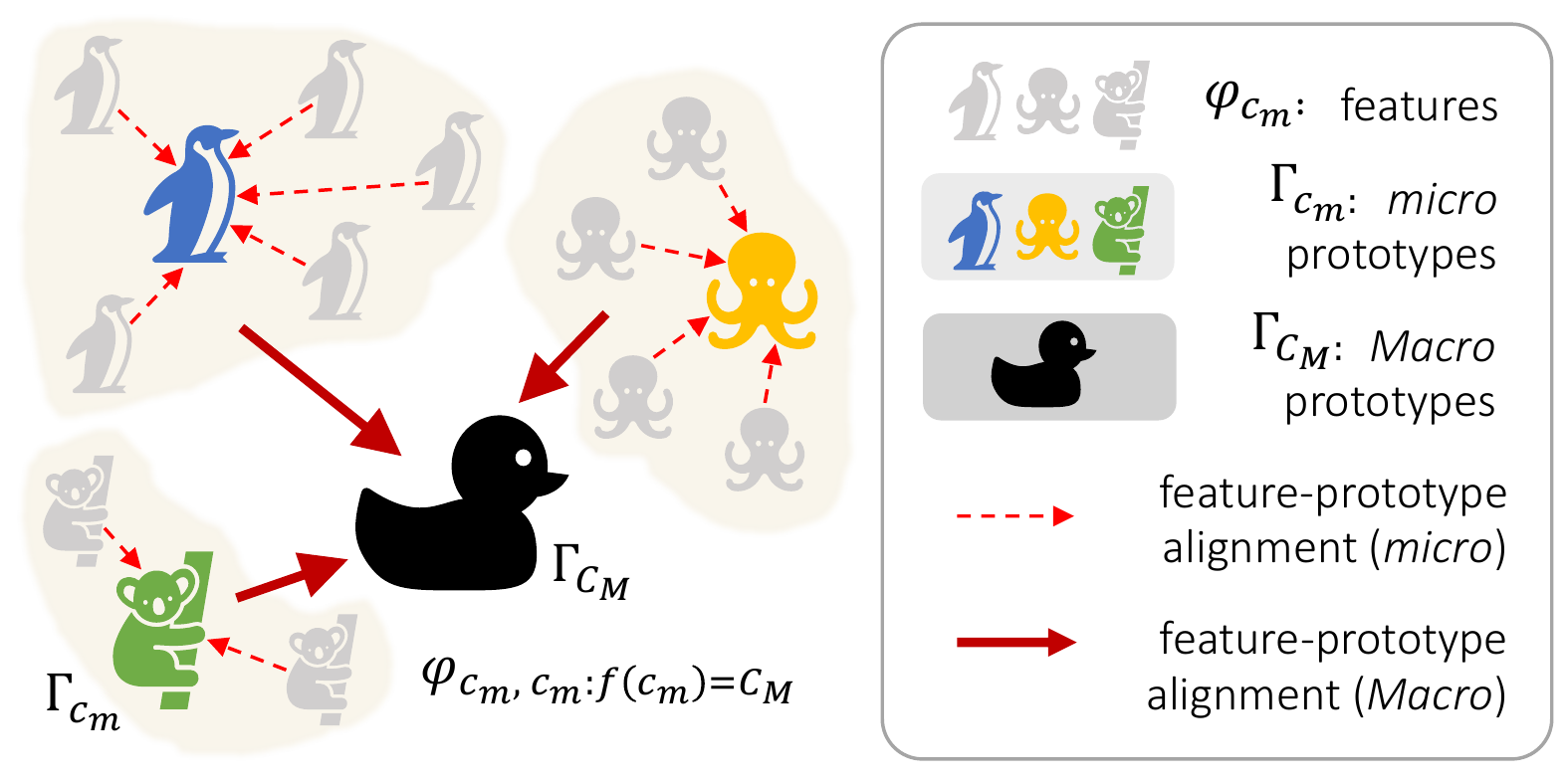}
     \vskip 1.5ex
 \caption{We identify semantic \textit{macro} communities (\eg, \textit{vehicles}) of \textit{micro} classes (\eg, \textit{car} and \textit{truck}) automatically analyzing the accuracy results of any semantic segmentation model. We regularize model training with $2$ components. Top: a macro-aware fairness ($\mathcal{F}$) score on the micro classes promotes homogeneous scores within each macro cluster. Bottom: class-conditional latent features-to-prototype alignment at $2$ levels (micro and macro) improves class-wise features discrimination.}
 \vskip -1.5ex
 \label{fig:graph_abs}
\end{figure}

\IEEEPARstart{S}{emantic} scene understanding is a challenging computer vision problem that finds application in various fields including autonomous driving, robot sensing, and virtual reality. 

Specifically, semantic segmentation is the most fine-grained scene understanding task that provides point-wise labeling on image pixels or 3D points. \sm{Since the refinement of classification accuracy can bring immeasurable benefits on different tasks (from navigation control to action planning), recent research has focused on improving different deep learning models for heterogeneous types of sensed data (\eg, RGB images, depth maps, and point clouds).}
\sm{Such improvements can be achieved in different ways. One possibility involves the adoption of enhanced processing architectures \cite{long2015fully} that better parameterize the input data with respect to the segmentation task. In this case, it is possible to improve the results sensibly at the cost of higher computational loads and memory requirements.}
Current state-of-the-art solutions are typically built on  the top of autoencoder architectures or fully-convolutional models \cite{long2015fully}, and their inner structure strongly depends on the task and the properties of the processed data, \rev{without relying on specific class-conditional constraints or abstractions.} 

\sm{A possible alternative consists in designing more accurate learning paradigms that are able to self-optimize the final performance even if architecture and datasets remain the same (without any additional intervention or information by the designer).}
To this extent, recent works \cite{li2020improving,hoyer2021three,zhu2020improving} have focused on \sm{the adopted} learning paradigms that maximize the semantic segmentation performance without increasing the size or complexity of the network. Such approaches exploit either body-edge features \cite{li2020improving}, self-supervised depth estimation \cite{hoyer2021three}, or pseudo labels \cite{zhu2020improving}, and do not perform any adaptive self-regularization estimated from a preliminary class-conditional accuracy.
\sm{The main advantage of learning-based enhancement is that the training process structure the data in a self-supervised manner, in such a way that the memory and computational requirements remain the same at inference time, but the final accuracy improves. This proves to be extremely desirable in many application scenarios (\eg, embedded deep learning scheme with real-time constraints) where low latency and a limited network size can not be bargained with improved performance.}

\sm{Following this trend, we propose LEAK (LEArning from mistaKes), a novel coarse-to-fine learning strategy that automatically optimizes the performance of a semantic segmentation network by shaping class subspaces according to classification errors and unbalancing. }
Remarkably, our approach does not imply additional supervision by human operators, making the full training process self-organized. 
\sm{The core idea relies on dividing the feature space in micro and \textit{macro} regions (where \textit{macro} space is an aggregation of \textit{micro} class subspaces fused according to their mutual misclassification probability) and balance them according to their representativeness.}

First, we use a pre-trained standard segmentation model to derive a confusion matrix over the (\textit{micro}) classes contained in our dataset. {Then, the optimization routine identifies \textit{macro} classes that include visually similar \textit{micro} classes by means of spectral clustering \cite{pothen89sc} on the confusion matrix.} 
{Empirically, we verified that \textit{macro} classes include similar semantic content, as expected.}
 
Such \textit{micro-macro} partitioning is used to derive different regularization terms. 
A fairness-enforcing constraint is included in the training loss to make classification errors uniformly distributed regardless of the sample frequency or accuracy per class (upper part of Fig.~\ref{fig:graph_abs}), and hierarchy-aware class-conditional regularization constraints are introduced to embed feature vectors of the same class tightly around their micro and macro prototypical representations (lower part of Fig.~\ref{fig:graph_abs}).

{We tested LEAK on} different point cloud semantic segmentation networks (RandLA-Net \cite{hu2020randla}, Cylinder3D \cite{zhou2020cylinder3d}, {RangeNet++ \cite{milioto2019rangenet++}}) and datasets, including sequential LiDAR data (SemanticKITTI \cite{behley2019semantickitti}) 
and static datasets (Semantic3D \cite{hackel2017isprs}, S3DIS \cite{armeni20163d}) acquired with laser scanners and other technologies. The proposed framework can be seamlessly adapted to different scenarios and is agnostic to the architecture and dataset. Its generality was also verified by adapting it to a standard semantic image segmentation dataset, \ie, Pascal-VOC2012 \cite{everingham2010pascal}, using DeepLabV3 \cite{chen2017rethinking}.

Some recent approaches \cite{zhou2022rethinking,toldo2021unsupervised,michieli2021continual,pinheiro2018unsupervised} have highlighted the opportunities of a feature-level clustering using class prototypes to characterize {a generic feature for each class; however, experimental results show that hierarchy-awareness can significantly boost semantic recognition.}

\sm{We can summarize the main contributions of this work as follows.
(1) We propose a general framework for semantic segmentation, adaptable to different experimental scenarios. In this way, LEAK proves to be very general and adaptable to different contexts.
(2) We identify a semantically-consistent partition of classes in \textit{macro} categories 
by inspecting the confusion scores of a standard model via spectral clustering. Previous coarse-to-fine approaches \cite{mel2020incremental,stretcu2020coarse,stretcu2021coarse,shenaj2022continual} usually need human supervision in defining the hierarchical split. LEAK adopts a fully automated procedure to extract and concretely encode such information, constraining the network with output-level fairness and feature-level regularization.
(3) We devise a hierarchy-aware fairness constraint to balance classification scores regardless of the frequency or accuracy of each class. Output-level fairness has been widely investigated in resource allocation \cite{jain1999throughput}; however, no prior {works include fairness measures in loss definitions for training deep architectures.}
(4) We compute a class-conditional hierarchical prototype structure that enforces an alignment of the generated feature vectors around their prototypical representation.
Some recent approaches \cite{zhou2022rethinking,toldo2021unsupervised,michieli2021continual,pinheiro2018unsupervised} have highlighted the opportunities of a feature-level clustering to characterize a generic feature for each class. However, experimental results show that hierarchy awareness can significantly boost semantic recognition.
(5) We benchmark our approach on different standard point cloud and RGB semantic segmentation datasets outperforming state-of-the-art architectures.}

\section{Related Work}\label{sec:related}

\textbf{Point Cloud Semantic Segmentation (PCSS)} has been tackled using different methods and architectures \cite{camuffo2022recent}. 
A first set of approaches relies on discretization methods that transform point clouds into discrete data structures. These structures can be dense, like voxels \cite{tchapmi2017segcloud} or octrees \cite{riegler2017octnet} or sparse, like permutohedral lattices \cite{rosu2021latticenet} and can be treated as three-dimensional images where convolutions can be applied.
Another category of methods consists in projecting the point cloud on a bi-dimensional structure to infer predictions and map it back in a later stage.
The projection methods are based either on multi-view \cite{su2015multi}, spherical \cite{milioto2019rangenet++} or cylindrical \cite{zhang2020polarnet} projections. 
Deep learning architectures used in these cases are usually well-established convolutional neural networks (CNN) pre-trained on image datasets.
Compared with discretization-based models, these methods are able to improve the performance for different tasks by taking multiple views of the object or scene of interest. 
In addition they are efficient in terms of computational complexity.
Finally, point-based methods avoid limitations posed by both projection- and discretization- based methods, \eg, loss of structural information, via direct processing of the raw point cloud data.
Among these methods we can distinguish point-wise MLP approaches \cite{qi2017pointnet,qi2017pointnet++,hu2020randla}, point-convolutions \cite{li2018pointcnn}, RNN-based \cite{huang2018recurrent,ye20183d} and graph-based methods \cite{wang2019dynamic,landrieu2018large}.

However, the most recent approaches rely on 4D convolutions \cite{fan2021pstnet} or \rev{transformers \cite{Guo_2021,lai2022stratified}} to accomplish the task and they need a huge computational power and storage capacity. More lightweight approaches consist in a mixture of methods; for example, some architectures provide voxel-wise predictions refined with point-wise labels \cite{zhou2020cylinder3d}. \rev{These approaches have been recently exploited with Knowledge Distillation methods that combine the two data representations in order to improve performance \cite{pvkd}.}
\\

\textbf{Prototype-based regularization} strategies have been successfully proposed to support deep learning architectures in solving a wide range of transfer learning problems, such as few-shot learning \cite{allen2019infinite,cermelli2020few,snell2017prototypical,dong2018few,chen2020compositional,yang2021part}, domain adaptation \cite{pan2019transferrable}, continual learning \cite{delange2021continual}, federated learning \cite{michieli2021prototype}. The common idea is to regularize model training by constraining the space of the extracted features.
Such techniques found applications in image classification \cite{Keswani_2022_CVPR,snell2017prototypical,pan2019transferrable,li2021prototypical}, image semantic segmentation \cite{zhou2022rethinking,dong2018few} and 3D point cloud segmentation \cite{zhao2021few,chen2020compositional}.

In \cite{snell2017prototypical} the idea that there exists an embedding in which points cluster around a single prototype representation for each class is first formulated, and then features are assigned to the class of the closest prototype (nearest neighbor). 
Since then, prototypes have been employed in many ways. Some works employ prototypical contrastive learning by clustering features of the same class tightly around their prototype, while spacing apart features of different classes \cite{li2021prototypical}. Matching of prototypes improved generalization across domains \cite{liu2021bapa,pan2019transferrable} and reduced forgetting when distilling knowledge from a support set of prototypes \cite{michieli2021prototype,cermelli2020few}.

Multiple class-conditional prototypical representations have been employed in \cite{zhao2021few,yang2021part,allen2019infinite} to better capture the complex statistical distribution of the extracted features. However, to the best of our knowledge, no prior work investigates the interaction of coarse- and fine-level prototypes to leverage standard supervised model training.
For point clouds, prototypes have been used in \cite{li2021prototypical} to support few-shot PCSS by either composing fine-level prototypes \cite{li2021prototypical} or by building an attention mechanism from multiple prototypes \cite{zhao2021few}.\\

\textbf{Hierarchical} composition of semantic representations has been explored in few previous work for part-based regularization \cite{li2021prototypical,yang2021part} and coarse-to-fine approaches where coarse-level classes are refined into finer categories \cite{mel2020incremental,stretcu2020coarse}.
However, they all require an explicit \textit{micro-to-macro} assignment.

\begin{figure*}[t]
    \centering
    \includegraphics[width=\textwidth]{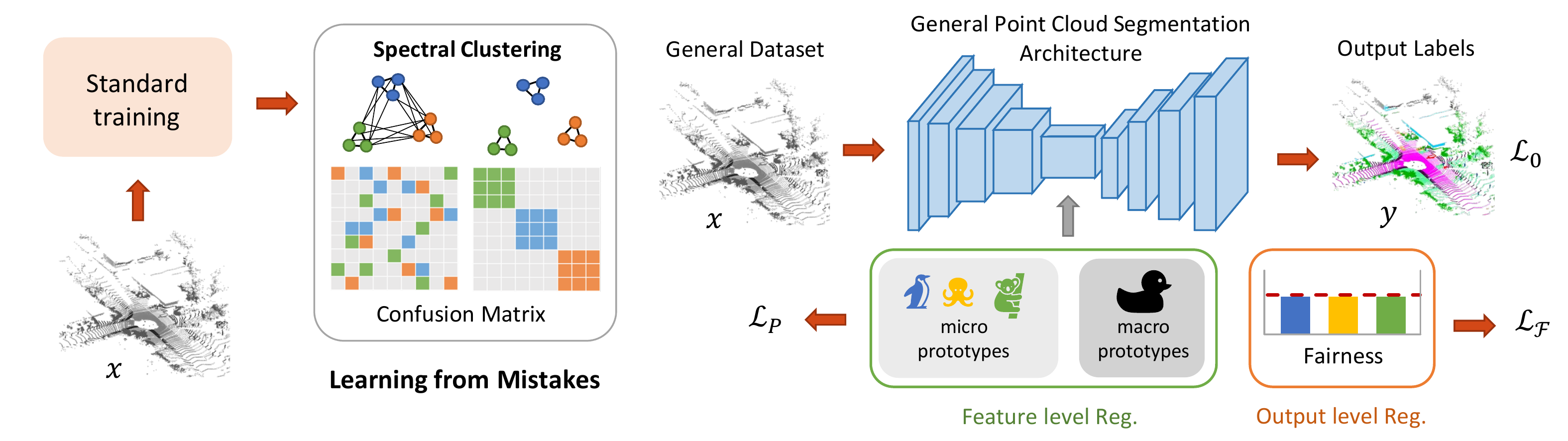}
    \caption{Overall pipeline of the proposed approach. First (left side), we analyze the results of a standard supervised learning performed by any off-the-shelf segmentation model identifying \textit{macro} communities of similar \textit{micro} semantic classes. Then (right side), we regularize the learning of the model by clustering features around their prototypical semantic representation at two levels (micro and macro) and by a macro-aware fairness score on the micro classes.}
    \label{fig:pipeline}
    \vskip -3ex
\end{figure*}

\section{Methodology}\label{sec:methods}
Given an input point cloud, \ie, a set of $N$  3D points $\rm X$ $ = \{ \bf x_1, x_2, \dots , x_N \} $, and a set of candidate semantic labels \noindent $\rm Y = \it \{y_1, y_2, \dots , y_N \}$, the objective of semantic segmentation is to associate each input point $\bf x_i$ with a semantic label $y_i$.

Such input points can be treated in several ways: (1) they can be discretized as voxels, (2) they can be projected and treated as 2D images, (3) or they can be processed directly as they are. {We devise experiments on each of the three methodologies, addressing the main focus on the direct processing method.} 
In the following paragraphs we provide a detailed explanation of each component, \ie, clustering on the standard model mistakes, class balancing through fairness enforcement, and feature-prototype alignment in the latent space.
Note that the first two components are totally independent of the model, while the latter weakly depends on how the construction of prototypical features occurs within the architecture, which in turn depends on the processing method used for point clouds.

An overall scheme of our approach is shown in Fig.~\ref{fig:pipeline}. Spectral clustering is applied on the confusion matrix inferred from the standard pre-trained model. The hierarchical partition obtained over the set of classes is used within the fairness objective at the output and to build prototypes at both \textit{micro} and \textit{macro} levels. This way the model \textit{learns from mistakes} by adopting a semantic-driven self-regularization approach, and obtains an overall improvement against the standard solution.

\subsection{Learning from Mistakes}
The first building block of our LEAK is the effective core of the self-regularization strategy {based on mutual semantic misclassifications} that \textit{learns from mistakes}.
Generally, standard supervised approaches train models from scratch relying on an annotated training set; coarse-to-fine hierarchical approaches exploit additional information, grouping ground truth labels \textit{a priori} into several \textit{macro} categories \cite{stretcu2020coarse}, generated via a human annotation activity.
Conversely, LEAK performs \textit{a posteriori} unsupervised clustering of classes, independently from the specific dataset and architecture.
Indeed, the class partition is derived from the misclassifications produced by the standard segmentation method. Such errors provide meaningful feedback about the feature space organization and allow highlighting of the classes that can be easily confused. Therefore it permits an optimization of their hidden representations enhancing the separation between the corresponding semantic regions.

A pre-trained standard segmentation model is employed to infer predictions {on the validation set}, computing the confusion matrix over classes. This matrix $\mathcal{A}$ is considered as an adjacency matrix associated with a complete graph network $\mathcal{G}$, where the different classes are assigned to nodes and the conditional error probabilities are the edge weights. We identify the nodes of $\mathcal{G}$ with $\{c_i\}, i \in [0, m)$, where $m$ is the total number of classes, and the edges with $\{d_{i,j}\}, i,j \in [0, m)$, where $i,j$ denote the ground truth and predicted class index, respectively. The edge $d_{i,j}$ is associated with the probability of classifying ground truth class $c_i$ into predicted class $c_j$. 
We use this representation to draw the subdivision in communities with a clustering algorithm, identifying $M$ clusters, defined by $\{C_i\}, i \in [0, M)$.

Specifically, this partitioning is performed via \textit{spectral clustering}  commonly used to identify communities of nodes in a graph based on the edges connecting them. The adjacency matrix $\mathcal{A}$ is provided as an input and consists of a quantitative assessment of the relative similarity of each pair of points in the dataset.
The algorithm follows an iterative procedure that exploits the eigenvalues of the similarity matrix to conduct dimensionality reduction and progressively subdivides the network into two clusters until the optimal number of communities is reached. This number is estimated \textit{a priori} thanks to the graph conductance measure \cite{onclusterings2001} where the optimal number of clusters corresponds to the number of local minima.
The communities found at this step represent the macro-grouping of classes. The left side of Fig.~\ref{fig:pipeline} shows a visual representation of the effects brought by the spectral clustering algorithm on the graph and the confusion matrix. Each color corresponds to a set of nodes (\ie, \textit{micro} classes) that belong to the same cluster (\ie, \textit{macro} class). The tree structure of the left side in Fig.~\ref{fig:graph_abs} is derived bottom-up with this approach and shows the classes' hierarchical organization.

\subsection{Feature-Prototype Alignment}
Prototypes (\ie, class centroids) are non-learnable vectors in the feature space that are representative of each semantic category that appears in the dataset \cite{zhou2022rethinking,NIPS2017_7a98af17,soundstream}. During training, the features extracted by the encoder contribute in forming the latent prototypical representation both for micro and macro classes. Class prototypes $\boldsymbol{\Gamma}_c$ ideally represent the features' objective for the respective class at each training step.

Their computation occurs in place with a running average updated at each training step with supervision. At training step $t$ with batch $\mathcal{B}$ of $B$ total samples, the prototypes are updated for a generic class $c$ as:
\begin{equation}\label{eq:proto}
    \boldsymbol{\Gamma}_{c}[t] = \frac{1}{{k}_{c}[t]} \Bigg( {k}_{c}[t-1] \cdot {\boldsymbol{\Gamma}_{c}[t-1] + n_c \sum_{\Hat{\boldsymbol{\varphi}}_c \in \mathcal{B}}{\Hat{\boldsymbol{\varphi}}_c}} \Bigg)
\end{equation}
where $\Hat{\boldsymbol{\varphi}}_c$ is a feature vector in the current batch $\mathcal{B}$ corresponding to class $c$, $k_c[t]$ is the number of feature vectors corresponding to class $c$ met in all previous batches, and ${n}_{c}$ is the number of feature vectors corresponding to class $c$ in current batch $\mathcal{B}$. 
Therefore, $k_c[t]=k_c[t-1]+n_c$ with $k_c[0]=0$.

\rev{The correspondence of each feature vector $\Hat{\boldsymbol{\varphi}}$ with class $c$ is based on the idea that a general encoder network preserves local structures from the input space, whether it is composed of convolutional or MLPs' layers. Therefore, the ground truth labels of each point cloud are tracked throughout the encoder to reach the latent space and provide semantic labels for it. Then, features with the same semantic class $c$ are aggregated to contribute in the construction of prototype $\boldsymbol{\Gamma}_c$.}

Class prototypes are initialized to $\boldsymbol{\Gamma}_c[0]=0$ $\forall c \in [0,m)$.

We use the $l_1$ norm $|| \cdot ||_1$ as metric distance. 
\rev{We report in \textit{Suppl. Mat.} an ablation on the loss function used.}
The specific loss function is defined as:
\begin{equation}\label{eq:protoloss}
    \mathcal{L}_{P_m} = \frac{1}{m}\sum_{c=0}^{m-1}\frac{1}{n_c}\sum_{\hat{\boldsymbol{\varphi}}_c \in \mathcal{B}}{||\boldsymbol{\Gamma}_c - {\hat{\boldsymbol{\varphi}}_c}||_{1}}.
\end{equation}

The integration of prototypical representations in the objective function promotes a self-driven progressive regularization of the latent space, forcing an alignment of new incoming features to their prototypes (lower of Fig.~\ref{fig:graph_abs}). We compute prototypes both at micro and macro levels. 

\rev{\textit{Macro} prototypes are obtained following Eq.~\eqref{eq:proto}, but considering the \textit{macro} class $C = f(c)$ instead of the \textit{micro} class $c$, with $f(\cdot)$ being the \textit{micro-to-macro} mapping function identified by our spectral clustering algorithm. 
The respective loss function $\mathcal{L}_{P_M}$ is drawn as in Eq.~\eqref{eq:protoloss}, considering $M$ \textit{macro} classes indexed by $C$.}

Including both micro and macro prototypes, we increase simultaneously coarse and fine expressiveness (division in well-separated clusters) for the latent representations, promoting a meaningful hierarchical organization of feature samples.
The addition of feature-level regularization has shown to be beneficial also for different tasks such as image semantic segmentation \cite{Xie2018LearningSR}.

{Note that features are latent representations of input points, but they have usually lower resolution and have different shapes with respect to the input tensors. To assign feature labels we must account for the specific encoder structure and sub-sampling method. Therefore, feature labels are evaluated by propagating input labels through the encoder.}
{The three selected architectures represent exemplar networks of the three most common methods to process point clouds: RandLA-Net \cite{hu2020randla} is a point-based method built of MLP layers and sub-samples points using random sampling; Cylinder3D \cite{zhou2020cylinder3d} partitions the 3D space discretizing point clouds with cylindrical voxel-like structures; RangeNet++ \cite{milioto2019rangenet++} projects point clouds on 2D surfaces to process them like images.}

\subsection{Attentive Fair Weighting}
To enforce the regularization effect of feature-prototype alignment, an attentive per-class weighting scheme is introduced.
{This constraint is derived from the experimental observation that the number of points per class plays a significant influence on the classification accuracy for that class. For example, in SemanticKITTI \cite{behley2019semantickitti} the most frequent classes, \eg, \textit{vegetation} or \textit{road}, obtain higher accuracy with respect to the least frequent ones, \eg, \textit{person}, independently on the underlying architecture.  Besides, in many practical applications, the least frequent classes are the most critical ones (\eg, \textit{person} in an automotive scenario).}

We propose a regularization objective derived from the Jain's fairness index $\mathcal{F}$ \cite{jain1999throughput} to provide a balanced per-class weighting. In other words, we address a resource allocation problem within each \textit{macro} class, considering \textit{micro} classes belonging to the same \textit{macro} class as the users that are sharing the same resource:
\begin{equation}
\mathcal{F}=\sum_{C=0}^{M-1} \frac{\big(\sum_{\boldsymbol{\pi}_c \in \mathcal{B}, c : f(c) = C}{\pi}_{c,c}\big)^2}{m_C \cdot\sum_{\boldsymbol{\pi}_c \in \mathcal{B}, c : f(c) = C}{\pi}_{c,c}^2}
\end{equation}
where $m_C$ is the number of (\textit{micro}) classes within \textit{macro} class $C$, ${\pi}_{c,c}$ represents the $c$-th element in vector $\boldsymbol{\pi}_c$, and $\boldsymbol{\pi}_c$ is the average prediction vector for class $c$, obtained as:
\begin{equation}
    \boldsymbol{\pi}_c = \frac{1}{\bar{n}_c} \sum_{\boldsymbol{p}_c \in \mathcal{B}}\boldsymbol{p}_c
\end{equation}
where $\boldsymbol{p}_c$ is a generic prediction vector with ground truth class $c$ and $\bar{n}_c$ is the number of points labeled as $c$ in the current batch $\mathcal{B}$.

A high fairness value denotes a truly balanced resource allocation among the entities, while low values of fairness show an unbalanced share of resources (upper part of Fig.~\ref{fig:graph_abs}). 

{Therefore, in order to preserve accuracy homogeneity among classes, we design a fairness-based loss function as follows:}
\begin{equation}
\mathcal{L}_{\mathcal{F}} = (1-\mathcal{F}).
\end{equation}

While prototype alignment forces a semantic target representation for each class at \textit{micro} and \textit{macro} levels, the attentive weighting constraint based on fairness aims at providing unbiased output predictions within each \textit{macro} class. In other words, by posing such weighting constraint to predictions of the same \textit{macro} class, we give the same importance to all the semantically consistent \textit{micro} classes.

\begin{table*}[t]
\centering 
\caption{Per-class IoU on SemanticKITTI \cite{behley2019semantickitti} dataset. $\dag$: model re-trained from the official codebase for a fair comparison. {\textbf{Bold} indicates best compared to baseline.}}
\label{tab:results:semantikitti}  \vspace{0.2cm}
\setlength{\tabcolsep}{7.45pt}
\renewcommand{\arraystretch}{1.1}
\begin{tabular}{lcccccccccccccccccccc}
\toprule
 \hspace{9.2em} & \rotatebox{90}{\textbf{mIoU}} & \rotatebox{90}{road} & \rotatebox{90}{swalk} & \rotatebox{90}{parking} & \rotatebox{90}{ot-ground} & \rotatebox{90}{building} & \rotatebox{90}{car}  & \rotatebox{90}{bike} & \rotatebox{90}{mbike} & \rotatebox{90}{truck} & \rotatebox{90}{ot-vehicle} & \rotatebox{90}{vegetation} & \rotatebox{90}{trunk} & \rotatebox{90}{terrain} &
\rotatebox{90}{person} & \rotatebox{90}{bicyclist} & \rotatebox{90}{mcyclist} & \rotatebox{90}{fence} & \rotatebox{90}{pole} & \rotatebox{90}{t-sign} \\
\end{tabular} \vskip -2.7em
\setlength{\tabcolsep}{3.5pt}
\begin{tabular}{l|c|ccccccccccccccccccc}
  & \\
 & \\
\midrule
PointNet \cite{qi2017pointnet} & 14.6 & 61.6 & 35.7 & 15.8 & 1.4 & 41.4 & 46.3 & 0.1 & 1.3 & 0.3 & 0.8 & 31.0 & 4.6 & 17.6 & 0.2 & 0.2 & 0.0 & 12.9 & 2.4 & 3.7  \\
SPG \cite{landrieu2018large} & 17.4 & 45.0 & 28.5 & 0.6 & 0.6 & 64.3 & 49.3 & 0.1 & 0.2 & 0.2 & 0.8 & 48.9 & 27.2 & 24.6 & 0.3 & 2.7 & 0.1 & 20.8 & 15.9 & 0.8 \\
PointNet++ \cite{qi2017pointnet++} & 20.1 & 72.0 & 41.8 & 18.7 & 5.6 & 62.3 & 53.7 & 0.9 & 1.9 & 0.2 & 0.2 & 46.5 & 13.8 & 30.0 & 0.9 & 1.0 & 0.0 & 16.9 & 6.0 & 8.9 \\
TangentConv \cite{tatarchenko2018tangent} & 40.9 & 83.9 & 63.9 & 33.4 & 15.4 & 83.4 & 90.8 & 15.2 & 2.7 & 16.5 & 12.1 & 79.5 & 49.3 & 58.1 & 23.0 & 28.4 & 8.1 & 49.0 & 35.8 & 28.5 \\
DarkNet21Seg \cite{behley2019semantickitti} & 47.4 & 91.4 & 74.0 & 57.0 & 26.4 & 81.9 & 85.4 & 18.6 & 26.2 & 26.5 & 15.6 & 77.6 & 48.4 & 63.6 & 31.8 & 33.6 & 4.0 & 52.3 & 36.0 & 50.0 \\
DarkNet53Seg \cite{behley2019semantickitti} & 49.9 & 91.8 & 74.6 & 64.8 & 27.9 & 84.1 & 86.4 & 25.5 & 24.5 & 32.7 & 22.6 & 78.3 & 50.1 & 64.0 & 36.2 & 33.6 & 4.7 & 55.0 & 38.9 & 52.2 \\
RangeNet53++ \cite{milioto2019rangenet++} & 52.2 & 91.8 & 75.2 & 65.0 & 27.8 & 87.4 & 91.4 & 25.7 & 25.7 & 34.4 & 23.0 & 80.5 & 55.1 & 64.6 & 38.3 & 38.8 & 4.8 & 58.6 & 47.9 & 55.9 \\
LatticeNet \cite{rosu2021latticenet} & 52.9 & 90.0 & 74.1 & 59.4 & 22.0 & 88.2  & 92.9 & 26.6 & 16.6 & 22.2 & 21.4 & 81.7  & 63.6 & 63.1 & 35.6 & 43.0 & 46.0 & 58.8 & 51.9 & 48.4 \\
ThickSeg \cite{gao2021thickseg} & 55.2 & 71.2 & 27.2 & 27.0 & 70.5 & 55.1 & 45.1 & 27.3 & 48.7 & 79.4 & 66.6 & 73.1 & 63.6 & 74.5 & 58.9 & 69.1 & 35.9 & 73.4 & 32.3 & 50.9 \\
FPS-Net \cite{xiao2021fps} & 57.1 & 91.1 & 74.6 & 61.9 & 26.0 & 87.4 & 91.1 & 37.1 & 48.6 & 37.8 & 30.0 & 80.9 & 61.2 & 65.0 & 60.5 & 57.8 & 7.5 & 57.4 & 49.9 & 59.2 \\
BAAF-Net \cite{qiu2021semantic} & 59.9 & 
90.9 & 74.4 & 62.2 & 23.6 & 89.8 & 95.4 & 48.7 & 31.8 & 35.5 & 46.7 & 82.7 & 63.4 & 67.9 & 49.5 & 55.7 & 53.0 & 60.8 & 53.7 & 52.0 \\
M-RangeSeg \cite{wang2022meta} & 61.0 & 93.9 & 50.1 & 43.8 & 43.9 & 43.2 & 63.7 & 53.1 & 18.7 & 90.6 & 64.3 & 74.6 & 29.2 & 91.1 & 64.7 & 82.6 & 65.5 & 65.5 & 56.3 & 64.2 \\
\midrule
RangeNet21\dag \cite{milioto2019rangenet++} & 46.1 & \textbf{93.7} & 81.0 & 43.3 & \textbf{0.0} & \textbf{83.3} & \textbf{93.7} & \textbf{23.4} & \textbf{38.8} & \textbf{19.9} & 12.2 & \textbf{81.8} & 46.0 & \textbf{72.4} & 27.3 & \textbf{49.9} & \textbf{0.0} & 48.8 & 36.6 & 28.3 \\
\textbf{{LEAK (RangeNet21\dag)}} & \textbf{46.4} & 93.6 & \textbf{81.2} & \textbf{46.9} & \textbf{0.0} & 82.8 & 90.8 & 22.0 & 37.8 & 8.7 & \textbf{16.6} & 81.5 & \textbf{46.7} & \textbf{72.4} & \textbf{28.8} & 49.7 & \textbf{0.0} & \textbf{52.5} & \textbf{39.5} & \textbf{29.4} \\ \midrule
RandLA-Net\dag \cite{hu2020randla} &  53.4 & \textbf{91.8} & 77.0 & \textbf{41.0} & 1.0 & \textbf{87.7} & \textbf{93.4} & 13.4 & 30.3 & 69.9 & 39.5 &  84.9 & \textbf{60.4} & 73.4 & 50.6 & 67.2 & \textbf{0.0} & 43.7 & 50.9 & 37.9 \\
\textbf{LEAK (RandLA-Net\dag)}  &  \textbf{54.7} & 91.4 & \textbf{77.2} & 39.9 & \textbf{1.6} & \textbf{87.7} & 93.3 & \textbf{17.5} & \textbf{32.4} & \textbf{78.0} & \textbf{42.0} & \textbf{85.3} & 58.0 & \textbf{73.9} & \textbf{54.2} & \textbf{69.6} & \textbf{0.0} & \textbf{43.9} & \textbf{52.8} & \textbf{40.5} \\
\midrule
Cylinder3D\dag \cite{zhou2020cylinder3d} & 64.7 & \textbf{94.4} & 81.1 & 49.0 & 0.3 & \textbf{89.5} & \textbf{96.9} & 39.1 & 64.3 & \textbf{87.3} & \textbf{63.9} & 87.4 & 69.0 & 72.7 & 74.8 & 90.6 & \textbf{0.1} & \textbf{55.6} & \textbf{64.4} & 49.2  \\ 
\textbf{LEAK (Cylinder3D\dag)}     & \textbf{65.2}   & 94.2 & \textbf{84.3} & \textbf{49.2} & \textbf{2.6} & 86.2 & 96.5 & \textbf{48.7} & \textbf{65.3} & 85.8 & 60.6 & \textbf{88.0} & \textbf{69.2} & \textbf{85.1} & \textbf{75.7} & \textbf{91.6} & 0.0 & 38.7 & 65.3 & \textbf{51.8}  \\ 
\bottomrule
\end{tabular}%
\end{table*}

\begin{figure*}[t]
    \centering
    \includegraphics[width=\textwidth]{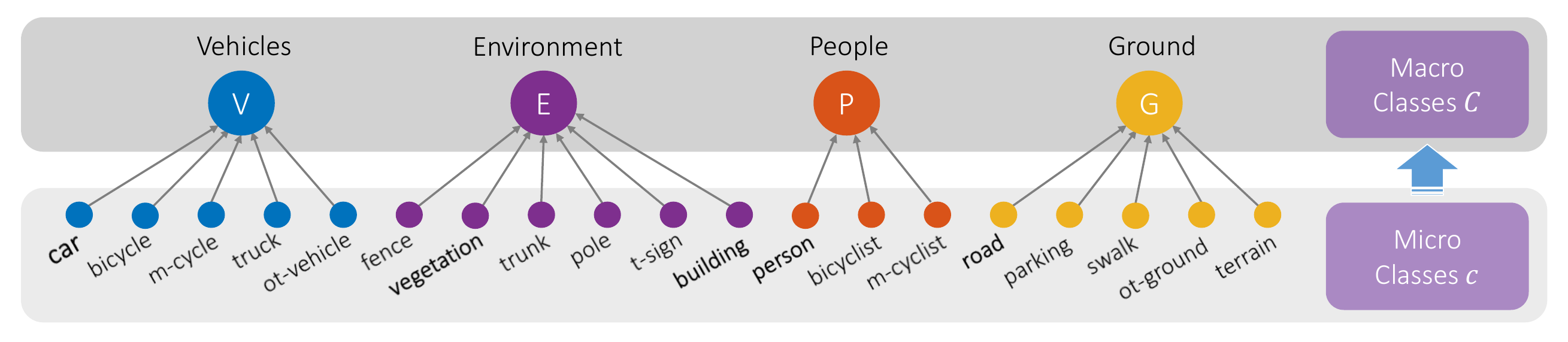}
    \caption{Hierarchical \textit{a posteriori} organization of SemanticKITTI \cite{behley2019semantickitti} classes.}
    \label{fig:hierarchy}
    \vskip -3ex
\end{figure*}

\subsection{Objective Function}
The training objective is given by the combination of the base loss function for each architecture ($\mathcal{L}_{0}$) with the additional objective given by the LEAK components. {The base loss function depends on the selected architecture. It corresponds to the standard cross-entropy loss with inverse class weighting for RandLA-Net \cite{hu2020randla} and RangeNet++ \cite{milioto2019rangenet++}, to the Lovasz-softmax loss \cite{berman2018lovasz} for voxel features plus the cross-entropy loss with inverse class weighting for point-feature refinement in Cylinder3D \cite{zhou2020cylinder3d}, and to the plain cross-entropy loss for DeepLabV3 \cite{chen2017rethinking}.}

The LEAK components are given by micro-level ($\mathcal{L}_{P_m}$) and macro-level ($\mathcal{L}_{P_M}$) feature-prototype alignment objectives, and a class-wise attentive weighting constraint ($\mathcal{L}_{\mathcal{F}}$). The full objective is then computed as:
\begin{equation}
    \mathcal{L}_{LEAK} = \mathcal{L}_{0} + \lambda_{P_m} \cdot \mathcal{L}_{P_m} + \lambda_{P_M} \cdot \mathcal{L}_{P_M} + \lambda_{\mathcal{F}} \cdot \mathcal{L}_{\mathcal{F}}
\end{equation}
where the balancing hyper-parameters have
been tuned using a validation set. 

\begin{figure*}[!ht]
\setlength{\tabcolsep}{0pt}
\def\arraystretch{0}
 \centering
 \begin{minipage}[c]{0.9\textwidth}
      \includegraphics[width=\textwidth]{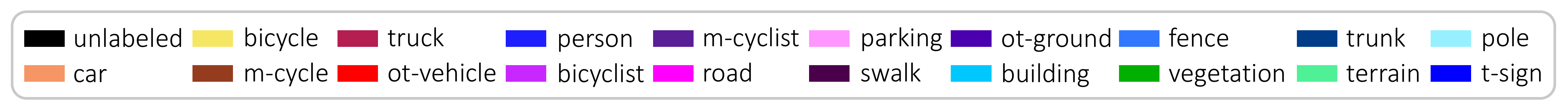}
      \label{fig:legend}
    \end{minipage}\hfill
  \begin{tabular}[c]{ccc}
     \begin{minipage}[c]{0.3\textwidth}
      \includegraphics[trim=0.3cm 0.5cm 0.3cm 1cm, clip,width=\textwidth]{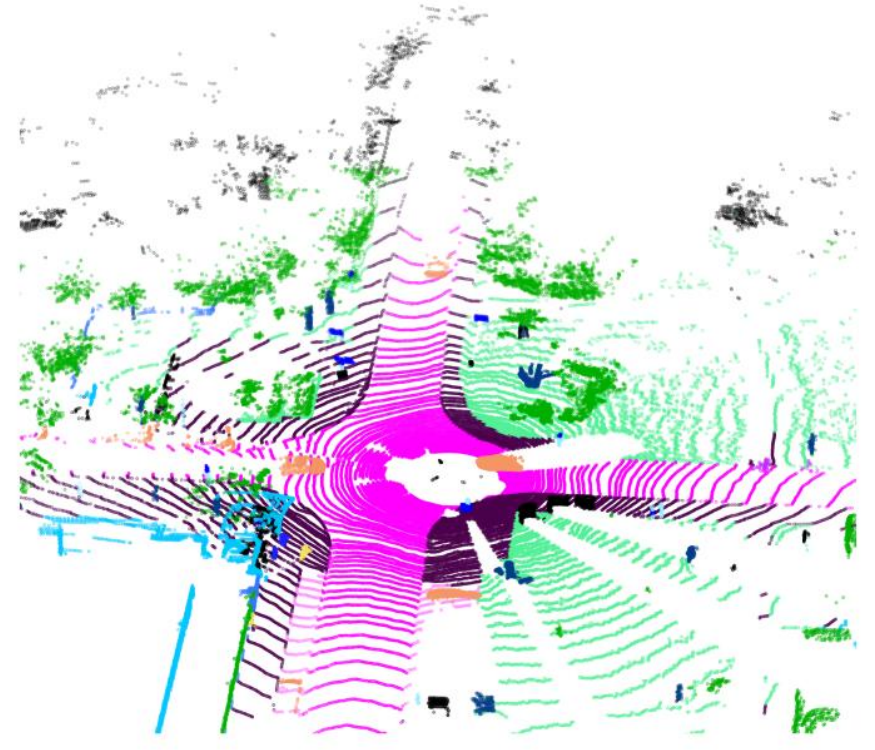}
      \label{fig:ceob}
    \end{minipage}&
   \begin{minipage}[c]{0.3\textwidth}
      \includegraphics[trim=0.3cm 0.5cm 0.3cm 1cm, clip,width=\textwidth]{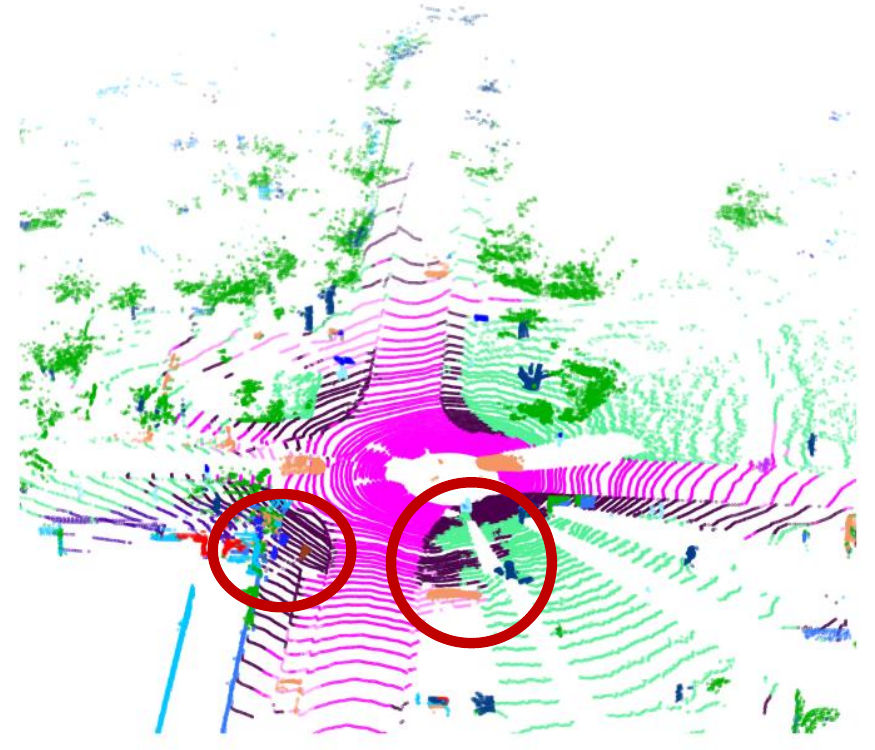}
      \label{fig:ceob}
    \end{minipage}&
    \begin{minipage}[c]{0.3\textwidth}
      \includegraphics[trim=0.3cm 0.5cm 0.3cm 1cm, clip,width=\textwidth]{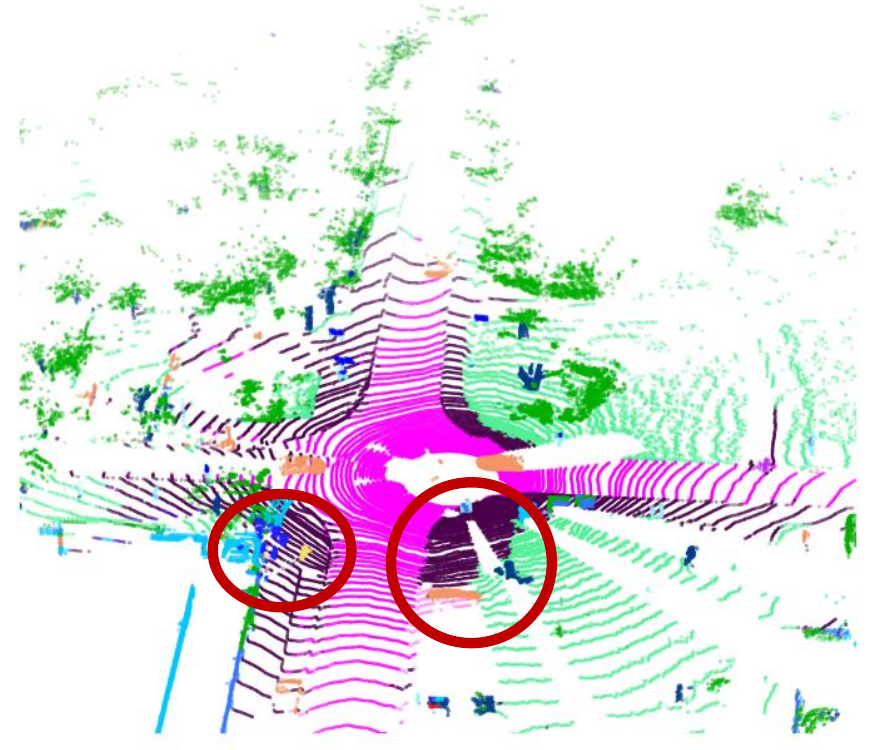}
      \label{fig:ceoc}
    \end{minipage}\\\hfill
        \begin{minipage}[c]{0.3\textwidth}
      \includegraphics[trim=0.3cm 0.3cm 0.3cm 1cm, clip,width=\textwidth]{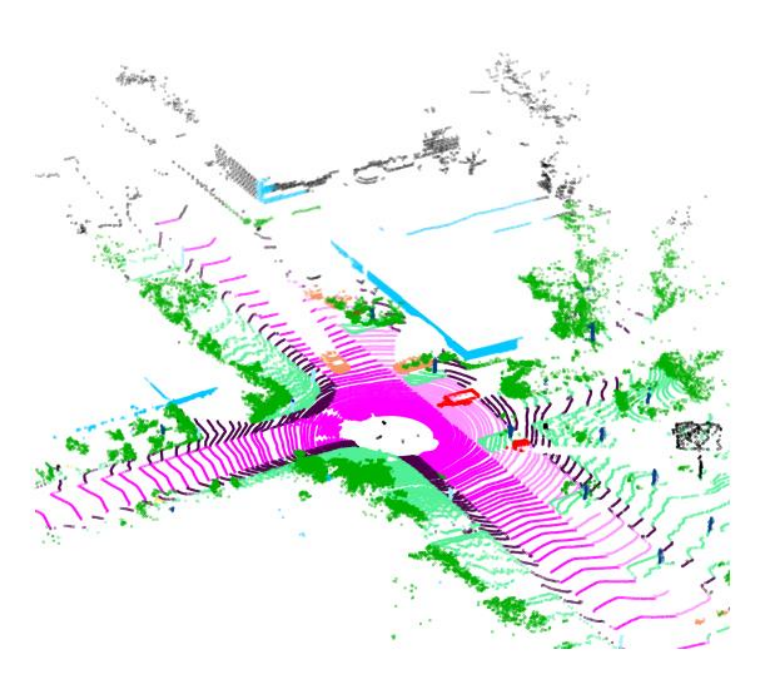}
      \vspace{-0.5cm}
      \caption*{Ground truth}
      \label{fig:caeob}
   \end{minipage}&
       \begin{minipage}[c]{0.3\textwidth}
      \includegraphics[trim=0.3cm 0.3cm 0.3cm 1cm, clip,width=\textwidth]{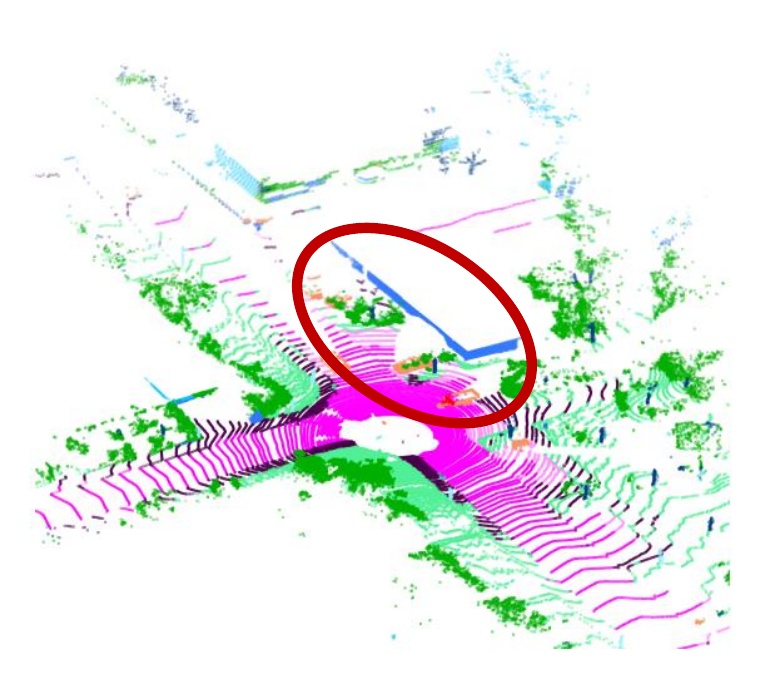}
      \vspace{-0.5cm}
      \caption*{Original}
     \label{fig:ceob}
    \end{minipage}&
    \begin{minipage}[c]{0.3\textwidth}
      \includegraphics[trim=0.3cm 0.3cm 0.3cm 1cm, clip,width=\textwidth]{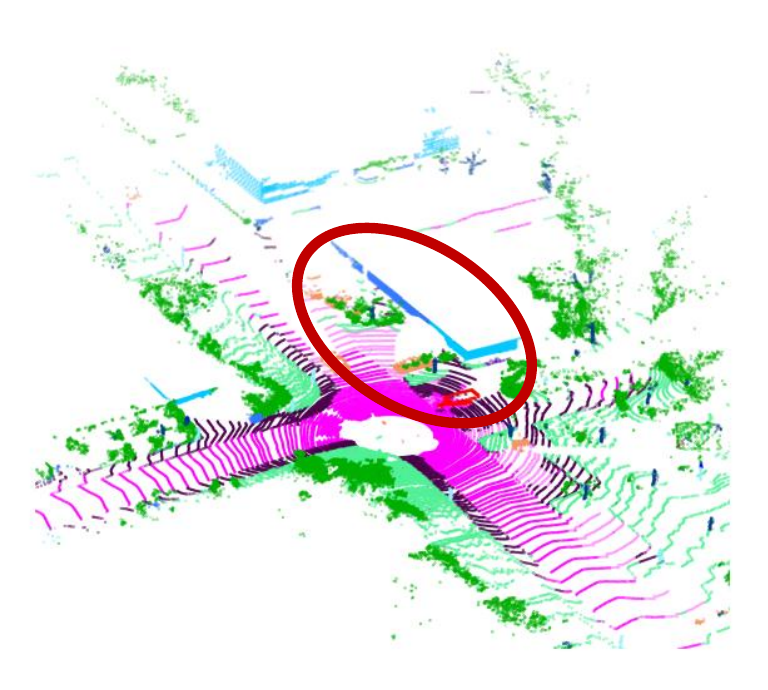}
      \vspace{-0.5cm}
      \caption*{LEAK}
      \label{fig:ceoac}
    \end{minipage}\\
  \end{tabular}    
  \caption{Qualitative results from SemanticKITTI \cite{behley2019semantickitti} with RandLA-Net \cite{hu2020randla}.}
  \label{fig:kittiqualitative}
\end{figure*}

\section{Training Procedure}\label{sec:training}

We experiment on publicly available benchmarks, using three point cloud datasets \sm{(2 outdoor and 1 indoor)} and one image dataset.

The \textbf{SemanticKITTI} \cite{behley2019semantickitti} dataset consists of $43552$ densely annotated \sm{outdoor} LiDAR scans. The training split contains $19130$ scans, while the validation split $4071$ scans (that we used for testing, as done by all competing works being the test labels not publicly available). The mean Intersection over Union (mIoU) score over $19$ categories is used as the standard metric and results are reported on the original validation set.

The \textbf{Semantic3D} \cite{hackel2017isprs} dataset consists of \sm{outdoor} static point clouds, $15$ for training and $15$ for online testing. Each point cloud has up to $108$ points. 
We only use color and spatial coordinates to train and test our models, following previous work \cite{hu2020randla}. We evaluate the performance via mIoU and OA on the $8$ classes.

The \textbf{S3DIS} \cite{armeni20163d} dataset consists of $271$ \sm{indoor} scans of medium-sized single rooms, with dense 3D points.
We use the standard $6$-fold cross-validation in our evaluation. We report mIoU, mean class Accuracy (mAcc), and Overall Accuracy (OA) of the $13$ classes.

{The \textbf{PascalVOC2012} \cite{everingham2010pascal} is used to validate our methods on \sm{RGB image samples}. It contains $10582$ images for training and $1449$ for online testing. We use the mIoU to evaluate the performance on the $21$ different classes.}

Our proposed strategy is agnostic to the backbone architecture. To prove it, in our experimental evaluation we use \textbf{RandLA-Net} \cite{hu2020randla}, \textbf{RangeNet++} \cite{milioto2019rangenet++},
\textbf{Cylinder3D} \cite{zhou2020cylinder3d} {and \textbf{DeeplabV3} \cite{chen2017rethinking} with their original hyper-parameters configuration}. 
We train RandLA-Net optimizing the network weights following \cite{hu2020randla}, with Adam optimizer and the same learning rate policy, momentum, and weight decay. The initial learning rate is set to $10^{-2}$, and decreased with a polynomial decay rule with power $0.95$. In each learning step, we train the models for $100$ epochs with a batch size of $6$ for SemanticKITTI and S3DIS, and $4$ for Semantic3D, as in the original model.
For Cylinder3D we train the model on SemanticKITTI using the standard configuration for $40$ epochs with batch size $2$ and learning rate $10^{-3}$. {Finally, for the evaluation on DeepLabV3 we follow the original configuration \cite{chen2017rethinking} and we trained the model for $30$ epochs with initial learning rate $7\times 10^{-4}$ and batch size $6$.}
We use an NVIDIA GeForce RTX 3090 GPU with CUDA 10.2 to train all the models.

\section{Experimental  Results}\label{sec:results}
We extensively evaluate the performance of our approach with different architectures and datasets, analyzing each component of LEAK separately. 

First of all, we show the results on SemanticKITTI \cite{behley2019semantickitti} dataset with two architectures: namely, RandLA-Net \cite{hu2020randla} and Cylinder3D \cite{zhou2020cylinder3d}. Then, we tested LEAK on two different datasets: S3DIS \cite{armeni20163d} and Semantic3D \cite{hackel2017isprs}.

The robustness of the method is further studied by applying LEAK in a different domain: we perform image semantic segmentation using DeepLabV3 \cite{chen2017rethinking} on PascalVOC 2012 \cite{everingham2010pascal} dataset.
Finally, the effect of each component is analyzed separately in an extensive ablation study.\\

\begin{table*}[t]
\begin{minipage}{0.56\linewidth}
\centering 
\caption{Quantitative results on the \revv{Semantic3D} \cite{hackel2017isprs} dataset with RandLA-Net \cite{hu2020randla}. \dag: model re-trained from the official codebase for a fair comparison. {\textbf{Bold} indicates best compared to baseline.}} 
      \label{tab:results:semantic3d}
\setlength{\tabcolsep}{7.45pt}
\renewcommand{\arraystretch}{1.1}
    \begin{tabular}{lcccccccccc}
        \toprule
        \hspace{9em} & \rotatebox{90}{\textbf{OA}} & \rotatebox{90}{\textbf{mIoU}} & \rotatebox{90}{mm-terrain} & \rotatebox{90}{nat-terrain} & \rotatebox{90}{h-vegetation} & \rotatebox{90}{l-vegetation} & \rotatebox{90}{buildings}  & \rotatebox{90}{hard scape} & \rotatebox{90}{s-artefacts} & \rotatebox{90}{cars} \\
                \end{tabular} \vskip -2.7ex \setlength{\tabcolsep}{3.5pt}
        \begin{tabular}{l|cc|cccccccc}
        & & \\
         \midrule
        SnapNet \cite{boulch2018snapnet} & 88.6 & 59.1 & 82.0 & 77.3 & 79.7 & 22.9 & 91.1 & 18.4 & 37.3 & 64.4 \\
        SEGCloud \cite{tchapmi2017segcloud} &  88.1 & 61.3 & 83.9 & 66.0 & 86.0 & 40.5 & 91.1 & 30.9 & 27.5 & 64.3 \\
        RF MSSF \cite{Thomas2018SemanticCO} & 90.3 & 62.7  & 87.6 & 80.3 & 81.8 & 36.4 & 92.2 & 24.1 & 42.6 & 56.6 \\
        MSDeepVoxNet \cite{roynard2018classification}  & 88.4 & 65.3 & 83.0 & 67.2 & 83.8 & 36.7 & 92.4 & 31.3 & 50.0 & 78.2 \\
        ShellNet \cite{zhang2019shellnet} & 93.2 & 69.3 & 96.3 & 90.4 & 83.9 & 41.0 & 94.2 & 34.7 & 43.9 & 70.2 \\
        GACNet \cite{Wang_2019_CVPR} & 91.9 & 70.8 & 86.4 & 77.7 & 88.5 & 60.6 & 94.2 & 37.3 & 43.5 & 77.8 \\
        SPG \cite{landrieu2018large} & 94.0 & 73.2 & 97.4 & 92.6 & 87.9 & 44.0 & 83.2 & 31.0 & 63.5 & 76.2 \\
        KPConv \cite{thomas2019kpconv} & 92.9 & 74.6 & 90.9 & 82.2 & 84.2 & 47.9 & 94.9 & 40.0 & 77.3 & 79.7 \\
        BAAF-Net \cite{qiu2021semantic} & 94.9 & 75.4 & 97.9 & 95.0 & 70.6 & 63.1 & 94.2 & 41.6 & 50.2 & 90.3 \\
    \midrule
        RandLA-Net\dag \cite{hu2020randla}   & 90.8   & 75.1 & 93.4 & 75.6 & 90.0 & \textbf{58.7} & \textbf{87.7} & 32.0 & \textbf{83.0} & 80.3 \\
\textbf{LEAK (RandLA-Net\dag)} & \textbf{90.9}  & \textbf{76.1} & \textbf{93.6} & \textbf{75.9} & \textbf{90.4} & 57.2 & 88.5 & \textbf{35.5} & 80.8 & \textbf{87.1}\\
\bottomrule
    \end{tabular}%
\end{minipage}\hfill
\begin{minipage}{0.4\linewidth}
	\centering \vspace{0.5cm}
    \includegraphics[trim=0.1cm 0cm 0cm 0cm, clip,width=\linewidth]{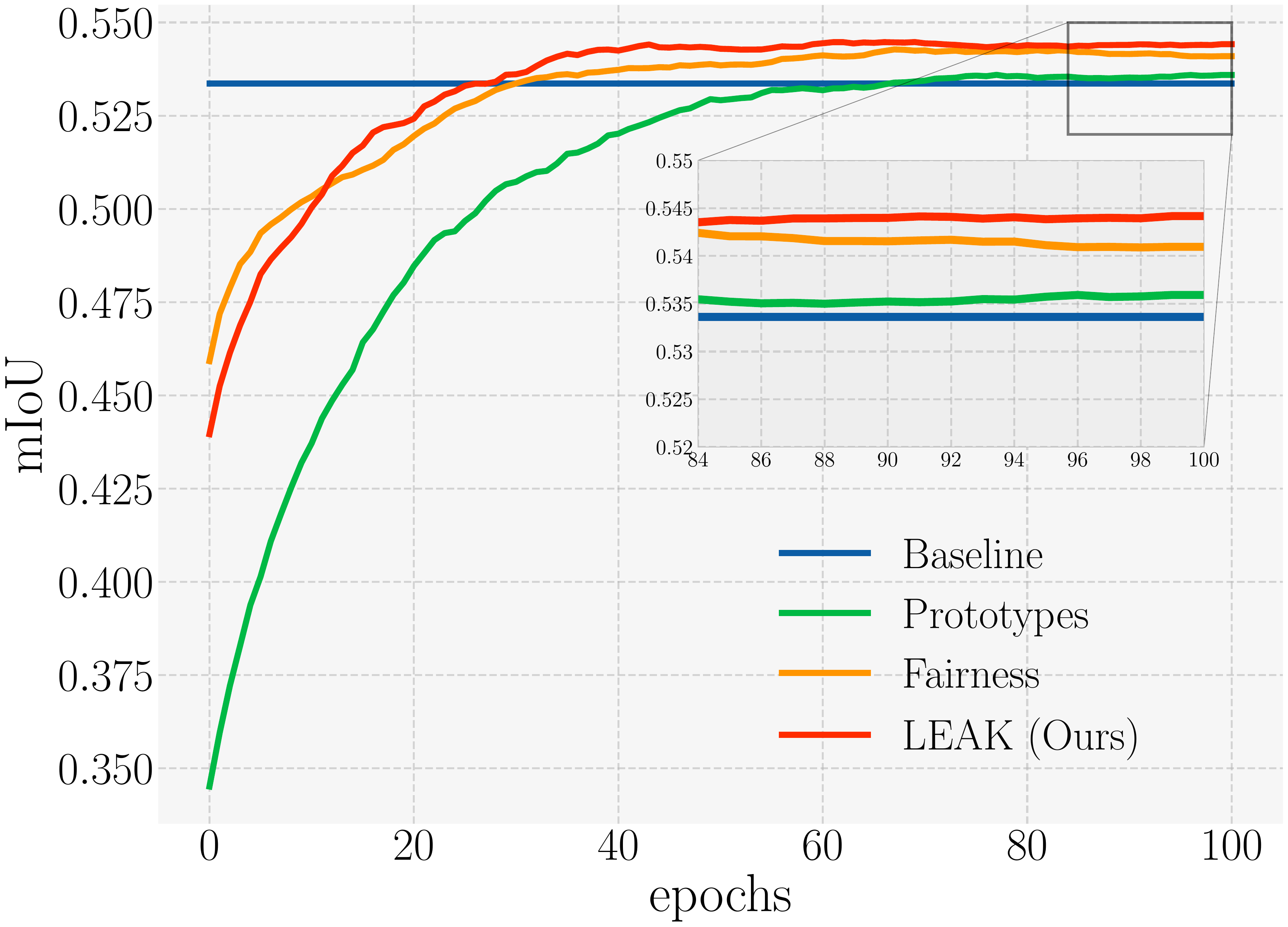}
	\captionof{figure}{mIoU curves comparing reference value (blue) to supervised training with the addition of prototype regularization (green), fairness (orange), or both (red). Curves smoothed via running average filter with window size $12$. LEAK provides higher mIoU and at the same time faster convergence speed.}
    \label{fig:miou}
\end{minipage}
\end{table*}

\textbf{SemanticKITTI.} We start our analyses on the public SemanticKITTI \cite{behley2019semantickitti} benchmark, reporting the results in Tab.~\ref{tab:results:semantikitti}, where our approach is compared against several well-established architectures.
 We took two among the most successful architectures and we applied LEAK during their model training.
The introduction of LEAK components improved the mIoU by $1.3\%$ on RandLA-Net \cite{hu2020randla} and by $0.7\%$ on Cylinder3D \cite{zhou2020cylinder3d}. 
{The performance drop in the retrained baselines are due to the use of a more recent version of CUDA library and other packages. More information about specific packages and versions are provided in the code repository.}

\begin{table}[t]
\centering \footnotesize
\caption{Quantitative results on the S3DIS \cite{armeni20163d} dataset (6-fold cross validation) with RandLA-Net \cite{hu2020randla}. \dag: model re-trained from the official codebase for a fair comparison.}
      \label{tab:results:s3dis}
\renewcommand{\arraystretch}{1.1}
\begin{tabular}{l|cccc}
\toprule
& \textbf{OA} & \textbf{mAcc} & \textbf{mIoU} \\
\midrule
PointNet \cite{qi2017pointnet} & 78.6 & 66.2 & 47.6 \\
PointNet++ \cite{qi2017pointnet++} & 81.0 & 67.1 & 54.5 \\
DGCNN \cite{wang2019dynamic} & 84.1 & | & 56.1 \\
SPG \cite{landrieu2018large} & 85.5 & 73.0 & 62.1 \\
DSPoint \cite{zhang2021dspoint} &
| & 70.9 & 63.3\\
PointCNN \cite{li2018pointcnn} & 88.1 & 75.6 & 65.4 \\
PointWeb \cite{zhao2019pointweb} & 87.3 & 76.2 & 66.7 \\
ShellNet \cite{zhang2019shellnet} & 87.1 & | & 66.8 \\
KPConv \cite{thomas2019kpconv} & | & 79.1 & 70.6 \\
JSNet \cite{zhao2020jsnet} & 71.7 & 88.7 & 61.7 \\
FKAConv \cite{boulch2020fka} & | & | & 68.4 \\
Point-PlaneNet \cite{peyghambarzadeh2020point} & | & 92.1 & 54.8 \\
JSENet \cite{hu2020jsenet} &  | & |& 67.7 \\
CT2 \cite{mazur2021cloudtransformers} & | & | & 67.4\\
\midrule
RandLA-Net\dag~\cite{hu2020randla} & 87.6 & 84.8 & 68.0 \\
\textbf{LEAK (RandLA-Net\dag)} & \textbf{88.1} & \textbf{85.4} & \textbf{68.5} \\
\bottomrule
\end{tabular}%
\end{table}

However, training with LEAK shows considerable improvements that outperform the baseline solution for both RandLA-Net and Cylinder3D. {Note that in RandLA-Net the prototypical representations are built based on the point-wise features, while in Cylinder3D on the voxel-wise ones.}

Looking at the per-class scores, it emerges how LEAK can provide more balanced results across the classes, thanks in particular to the fairness constraint.
The automatic hierarchical grouping for SemanticKITTI classes is shown in Fig.~\ref{fig:hierarchy}, where we can appreciate that each \textit{macro} class contains semantically similar \textit{micro} classes.
\revv{It is also possible to notice that the per-class IoU increases on the least populated classes (made exception for \emph{truck} in RangeNet21). In this case, fairness weighting combined to a higher misclassification probability depending on misleading object shapes after projection (RangeNet is based on a set of convolutional layers applied on images after spherical projections) induces a higher error probability on the points along the border of the object. This inconvenience is solved whenever applying LEAK to the other two architectures (RandLA-Net and Cylinder3D) that process array points in 3D coordinates.}

Fig.~\ref{fig:kittiqualitative} shows some qualitative results that compare the predictions of RandLA-Net trained with or without our LEAK. 
The most relevant improvements are highlighted with red circles. We observe that prediction errors of the original model often involve semantically and geometrically similar classes: \textit{sidewalk} is misclassified as \textit{road} in the first row, \textit{building} is misclassified as \textit{fence} in the second row. 
Our approach, instead, can correctly label these objects, thanks to the increased latent space self-regularization and fairness of accuracy across the same macro class.
Indeed, \textit{road} and \textit{sidewalk} belong to the same \textit{a posteriori} macro class, similarly to \textit{fence} and \textit{building}. Further qualitative results are provided in \textit{Suppl.\ Mat.}\\

\rev{Moreover, to verify the robustness of LEAK, we perform additional experiments on two other datasets, with different properties and acquired with different methodologies with respect to SemanticKITTI. We select S3DIS \cite{armeni20163d} and Semantic3D \cite{hackel2017isprs} datasets, employing RandLA-Net.
The strong generalization ability of our approach emerges clearly, despite the huge gap in the nature of the considered datasets. Indeed, the three point cloud datasets differ greatly in terms of the number of scenes, the number of points per scene, the acquisition methodologies, the environment (indoor/outdoor), and the number of classes. {SemanticKITTI scenes are organized in sequences and distributed according to sparse concentric regions. Instead, static datasets present denser and regularly distributed point regions.}
Qualitative results are provided in \textit{Suppl.\ Mat}.\\

\textbf{S3DIS.}
Results on S3DIS are shown in Tab.~\ref{tab:results:s3dis}.
RandLA-Net with LEAK achieves remarkable results and, in particular, outperforms the baseline training scheme by $0.5\%$ of mIoU. The improvement is shown also according to the other metrics, with an increase of $0.5\%$ and $0.8\%$ for the OA and mAcc, respectively.\\

\textbf{Semantic3D.}
The per-class results in terms of mIoU and OA achieved on Semantic3D are reported in Tab.~\ref{tab:results:semantic3d}. RandLA-Net with LEAK outperforms all the reported approaches, boosting the baseline of $1.0\%$ mIoU.}

\begin{table}[t]
\centering \footnotesize
\caption{Ablation study on SemanticKITTI \cite{behley2019semantickitti} dataset with RandLA-Net \cite{hu2020randla}. 
*: measure of the prediction coherence within the macro classes.}
\label{tab:ablation}
\setlength{\tabcolsep}{3.5pt}
\renewcommand{\arraystretch}{1.1}
\begin{tabular}{l|c|cccc|c|c|c}
\toprule
& $\mathcal{L}_0$ & $\mathcal{L}_{\mathcal{F}}$ & $\mathcal{L}_{P_M}$ & $\mathcal{L}_{P_m}$ & SC & \textbf{mIoU} & \textbf{{fwIoU}} & \textbf{{hIoU*}}\\
\midrule
Baseline & \checkmark & & & & & 53.4 & 81.4 & 77.9 \\
Prototypes & \checkmark & & \checkmark & \checkmark & & 53.9 & 81.5 & \textbf{85.2} \\
Fairness & \checkmark & \checkmark & & & & 54.4 & 82.0 & 84.8 \\ \midrule
\textbf{LEAK} &\checkmark & \checkmark & \checkmark &\checkmark &\checkmark &  \textbf{54.7} & \textbf{82.1} & 83.1 \\
\bottomrule
\end{tabular}%
\end{table}

\begin{table}[t]
    \centering
        \caption{\revv{Recent state-of-the-art approaches results in terms of mIoU for standard method and equipped with LEAK, which is shown to boost results on every type of architecture and dataset.}}
    \begin{tabular}{cc|cc}
    \toprule
        \textbf{Model} & \textbf{Dataset} & \textbf{Std.} & \textbf{LEAK}\\
        \midrule
        SphereFormer\cite{lai2023spherical} & SemanticKITTI \cite{behley2019semantickitti} & 67.8 & \textbf{68.5}\\
        PVKD \cite{pvkd}  & SemanticKITTI \cite{behley2019semantickitti} & 67.9 & \textbf{68.4}\\
        PVCNN \cite{liu2019pvcnn} & SemanticKITTI \cite{behley2019semantickitti} & 17.2 & \textbf{17.8}\\
        \midrule
        Strat.Transformer \cite{lai2022stratified} & S3DIS \cite{armeni20163d} & 66.4 & \textbf{66.9}\\
        Pt.Transformer \cite{zhao2021point} & S3DIS \cite{armeni20163d} & 70.4 &  \textbf{71.8}\\
        PVCNN \cite{liu2019pvcnn} & S3DIS \cite{armeni20163d} & 55.6 & \textbf{56.1}\\
    \bottomrule
    \end{tabular}
    \label{tab:resp}
\end{table}

\begin{table}[t]
    \centering
    \footnotesize
\caption{Ablation study on hierarchical grouping and weighting schemes using SemanticKITTI \cite{behley2019semantickitti} dataset. {*: coarse-to-fine is intended as a two-steps training, half of the total epochs on \textit{macro} classes and half on \textit{micro} classes, using LEAK class partitioning.}}
\label{tab:rebuttal} 
\setlength{\tabcolsep}{3.5pt}
\renewcommand{\arraystretch}{1.1}
    \begin{tabular}{l|c|c}
    \toprule
 Model & Modifications & \textbf{mIoU} \\
\midrule
RandLA-Net &  | &  53.4 \\
\textbf{LEAK (RandLA-Net)} &  | & \textbf{54.7} \\
\textbf{LEAK (RandLA-Net)} & \textit{a priori} grouping  &   54.2  \\
{RandLA-Net} & coarse-to-fine*      & 53.8    \\
RandLA-Net & weight scheme of SalsaNext \cite{cortinhal2020salsanext}  & 52.3 \\
\midrule
Cylinder3D & | & 64.7  \\ 
\textbf{LEAK (Cylinder3D)} & | & \textbf{65.2}  \\
\textbf{LEAK (Cylinder3D)} & \textit{a priori} grouping  &   64.2 \\
Cylinder3D & inverse frequency weighting &  64.8 \\ 
Cylinder3D & weight scheme of  SalsaNext \cite{cortinhal2020salsanext} &  63.5 \\ 
\bottomrule
\end{tabular}%
\end{table}

\section{Ablation Studies}
\label{subsec:ablation}
We report an extensive ablation study to fully validate our approach. First, we show task generalizability, reporting the results on VOC2012 \cite{everingham2010pascal}. Then, we devise  separate studies on each component of our LEAK.

\subsection{Task generalization.}
\textbf{VOC2012.}
We show the robustness of LEAK on segmentation of RGB images rather than point clouds.
 To perform image semantic segmentation we consider the VOC2012 \cite{everingham2010pascal} dataset and we use DeepLabV3 \cite{chen2017rethinking}, with ResNet-101 \cite{He2016DeepRL} as backbone (weights pre-trained on ImageNet \cite{Deng2009ImageNetAL}). 
Also in this case, LEAK brings an improvement compared to the standard method. In terms of mIoU, we obtain a value of $66.4\%$ using the standard approach and $66.7\%$ when LEAK is introduced.
Such a result further proves the validity of our method, showing that it is agnostic not only to the backbone architecture and the dataset but even, more generally, to the task. 

\revv{In addition, we report in Tab.~\ref{tab:resp} further experiments on recent architectures for Point Cloud Semantic Segmentation showing that LEAK improves baseline performance on top of every segmentation architecture, also in case transformer-based architectures are employed.}

\subsection{LEAK Components}
A first set of analyses were carried on the SemanticKITTI dataset with RandLA-Net architecture.
Tab.~\ref{tab:ablation} highlights the contribution of each component of LEAK on the final model scores. 
{We observe that both the fairness and the feature-prototypes alignment constraints significantly increase the final mIoU over the original model by $1\%$ or $0.5\%$ respectively. The combination of such regularization terms produces non-overlapping and mutual benefits, allowing for an increment of $1.3\%$ in the final mIoU. Experimental results show that a fundamental requirement for the success of our approach is the \textit{a posteriori} hierarchical clustering of the micro classes;} indeed, employing a semantic-unaware random clustering of micro classes leads to a small mIoU of $51.3\%$, which is even lower than the original approach ($-2.1\%$).

{Also, we report two additional metrics to show the effects of each component: the frequency weighted Intersection over Union (fwIoU) that weights each class importance according to their appearance frequency, and we introduce the hierarchical Intersection over Union (hIoU), defined as:
\begin{equation}
    \text{hIoU} = \frac{1}{M + 1} \sum_{i=0}^{M}\frac{q_{ii}}{\sum_{j=0}^{M}q_{ij} + \sum_{j=0}^{M} q_{ji} - q_{ii}}
\end{equation}
where $q=f(p)$, such that each prediction $p$ is mapped on the respective macro class with $f(\cdot)$. This metric is computed in such a way that every predicted micro class should be equal to the target macro class that contains it. Its purpose is to underline the weight of the hierarchical self-induced assessment over the original model. In fact, the disentangling effect of fairness and prototypes lead to an improvement of $6.9\%$ and $7.3\%$ respectively, in terms of hIoU, which is slightly reduced in the final model ($+5.2\%$). We can appreciate improved results also in the fwIoU ($+0.7\%$), where the main contribution is given by the class-balancing effect of fairness.}

Fig.~\ref{fig:miou} compares the temporal evolution of the mIoU score over different training epochs for each ablation model.
We observe that the introduction of feature-prototype alignment delays the increase in the mIoU compared to the original baseline model (green curve), but leads to an overall improvement of $0.5\%$ mIoU against the original model in the long term.
This delay is partially generated by the dominance of the new loss terms over the cross-entropy loss. In addition, this performance degradation proves initial prototypes to be quite unreliable, since they are computed just on few feature samples. 
As training proceeds, the prototypical features become progressively better defined and separated, leading to an increment in the overall accuracy. 

{Tab.~\ref{tab:rebuttal} reports further ablation results. We report tests using other cross-entropy weighting schemes (that in general helps class balancing), and tests using \textit{a priori} standard class grouping, as in \cite{behley2019semantickitti}. We extended the ablation on micro-macro grouping for both RandLA-Net and Cylinder3D, outperforming standard coarse-to-fine strategy and some other state-of-the-art weighting schemes, including SalsaNext \cite{cortinhal2020salsanext} (\ie, squared root of inverse class frequency), and inverse class frequency weighting on Cylinder3D.
Our accuracy gains are robust across setups, improving training with no architectural changes.}

\begin{figure}[t]
    \centering
    \begin{minipage}{0.24\textwidth}
    \centering
        \includegraphics[trim=1.5cm 0cm 0cm 3.5cm, clip, width=1.1\textwidth]{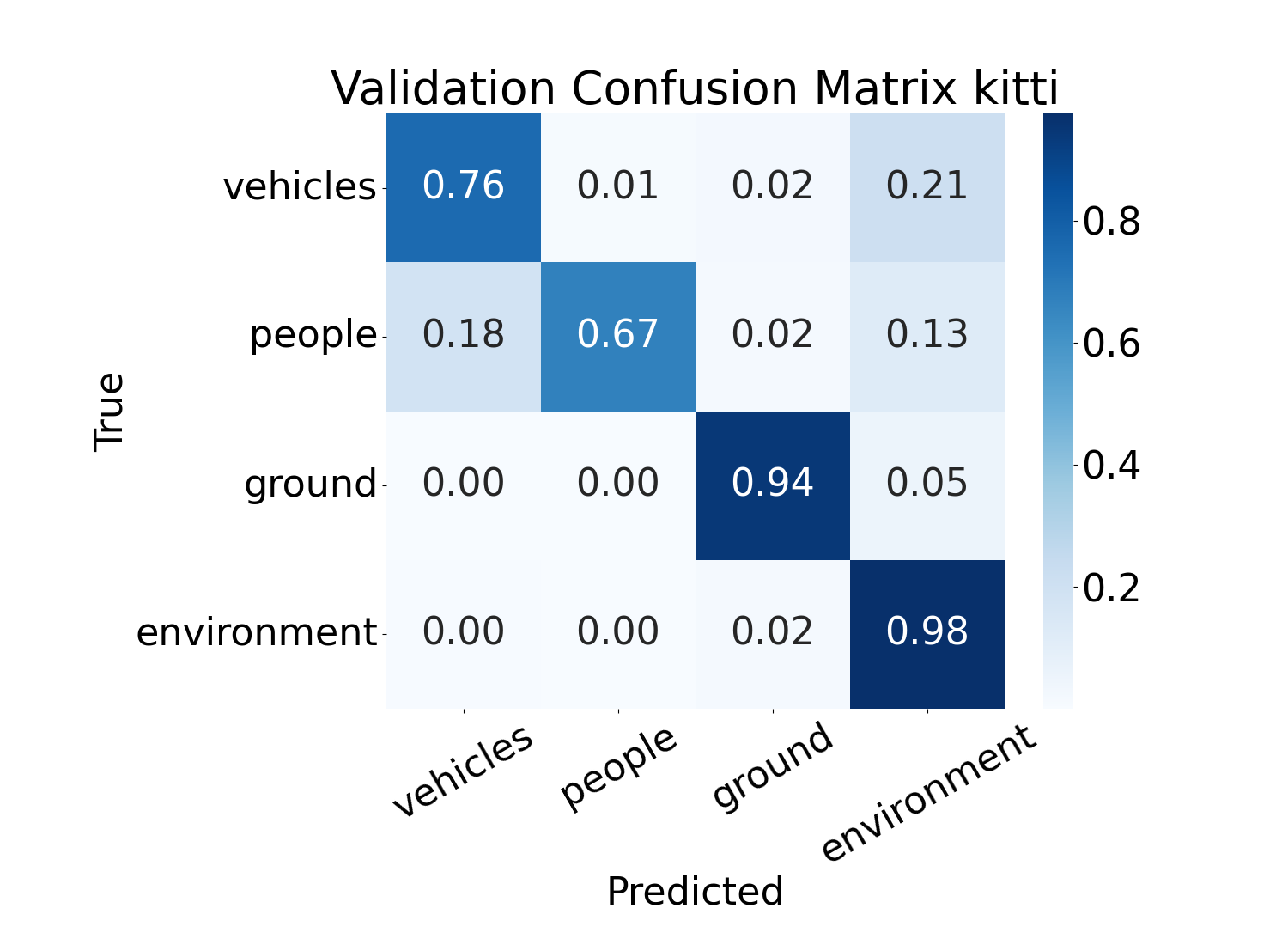}
    \caption*{Pre-LEAK}
    \end{minipage}
    \begin{minipage}{0.24\textwidth}
    \centering
    \includegraphics[trim=1.5cm 0cm 0cm 3.5cm, clip, width=1.1\textwidth]{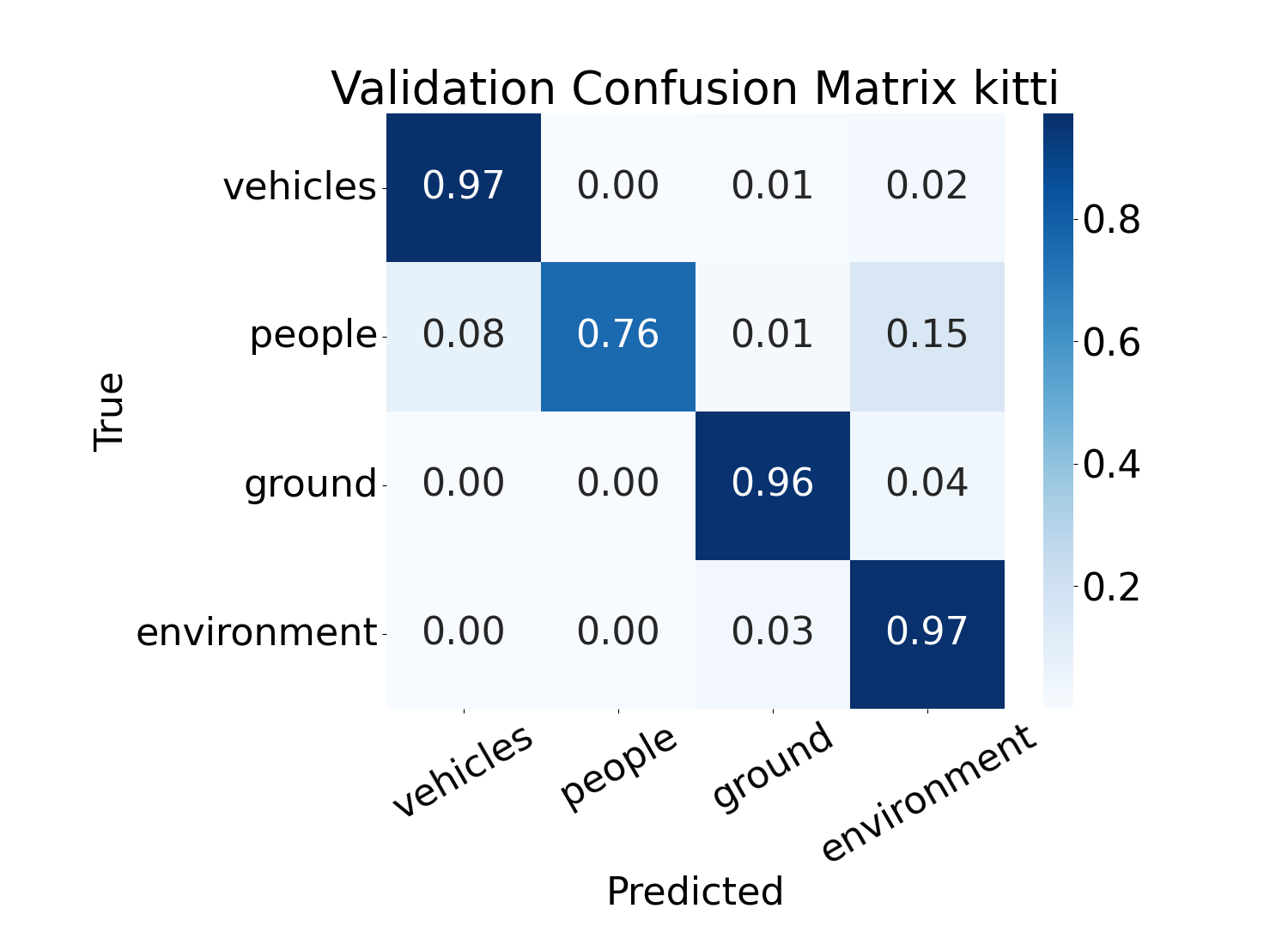}
     \caption*{Post-LEAK}
     \end{minipage}
    \caption{\rev{Confusion matrices of the fine-grained predictions grouped according to the \textit{macro} classes identified \textit{a posteriori} pre- and post- LEAK on RandLA-Net \cite{hu2020randla} with SemanticKITTI \cite{behley2019semantickitti}.}}
    \label{fig:confmat}
\end{figure}

\rev{Fig.~\ref{fig:confmat} reports the confusion matrices obtained with the standard method (Pre-LEAK) and the confusion matrix analyzed at the end of the training with LEAK (Post-LEAK). 

The overall {error percentages} are reduced and fall mostly in the \textit{macro} categories derived with \textit{a posteriori} clustering. 
In particular, we can notice that LEAK brings a great reduction in the misclassification of \textit{vehicles} with \textit{environment}, from $21\%$ misclassified samples to $2\%$. Also, we can see an improvement in the misclassification of \textit{people} with \textit{vehicles}, from $18\%$ to $8\%$, with the overall results on the diagonal either improving or remaining the same.}

\subsection{Feature-prototype alignment}

To measure the impact of feature-prototype alignment module we compute some metrics to compare LEAK against the original model, as shown in Tab.~\ref{tab:res:metrics}.
First, we define the \textit{inter-proto angle} $\Theta_{\Gamma}$ as the angle between prototypes in the n-dimensional space, \ie:
\begin{equation}
    \Theta_{{\Gamma}} = \frac{1}{m^2}\sum_{i \neq j; i,j\in [0,m)}{\frac{<\boldsymbol{\Gamma}_i, \boldsymbol{\Gamma}_j>}{||\boldsymbol{\Gamma}_i||\cdot||\boldsymbol{\Gamma}_j||}}
\end{equation}
in order to measure the relative distancing among prototypes.
Then, we compute the \textit{Class Center Distance} (CCD) as the average distance between features $\boldsymbol{\varphi}_c$ of class $c$ and the average feature array $\Bar{\boldsymbol{\varphi}}_c$ of class $c$ \cite{Wang2020ClassesMA} (computed at the end of training), \ie:
\begin{equation}
    \text{CCD} = \frac{1}{N}\sum_N \frac{1}{N_c} \sum_{c \in [0,m)}{||\boldsymbol{\varphi}_c - \Bar{\boldsymbol{\varphi}}_c||.}
\end{equation}
The parameter $N$ is the total number of samples in the dataset, while $N_c$ is the total number of samples for class $c$. The purpose of CCD is to parameterize how tightly the features are clustered around their class center.
Similarly, we define also the \textit{Prototype Distance} (PD) as the average distance of the features of class $c$ from prototype $\boldsymbol{\Gamma}_c$ (which is computed by a running average during the training phase), and the \textit{Proto-Center Distance} (PCD) as the distance between $\Bar{\boldsymbol{\varphi}_c}$ and $\boldsymbol{\Gamma}_c$.
The metric values are reported in Tab.~\ref{tab:res:metrics} and confirm that feature-prototype alignment thins the regions associated to a specific class as CCD and PD are much lower when feature-prototype alignment is enabled. Additionally, the increased value of $\Theta_\Gamma$ in LEAK with respect to the original training highlights that a progressive spacing is introduced among prototypes thus refining the class latent feature representations.

\subsection{Attentive fair weighting}
The effect of attentive fair weighting is instead to boost the performance from the immediate beginning of the training (green curve in Fig.~\ref{fig:miou}), obtaining faster convergence and overall higher precision. The attentive fair weighting curve in  Fig.~\ref{fig:miou} reaches $50\%$ mIoU in only $10$ epochs, \ie, the $91.4\%$ of the final score ($54.7\%$). 
We can explain this rapid increase in the mIoU with the re-weighting, computed at each training step, which forces an equalization among different classes. Fair weights computation is performed class-wise, and shows beneficial effects from the beginning, balancing some unfair weighting of the cross-entropy that privileges the most diffused classes.
In the final LEAK method, where all the regularization terms are combined, the improvement due to fair weighting is slightly attenuated by feature-prototypes alignment in the first $15$ epochs but performs the best in the long run, improving the mIoU of $1.3\%$ with respect to the standard configuration.
In the lower of Tab.~\ref{tab:res:metrics2} we appreciate that per-class accuracy are more balanced when fairness is enabled (see \textit{Suppl.\ Mat.}). The reported metrics are computed on the per-class results of Tab.~\ref{tab:results:semantikitti} and confirm that fair weighting balances accuracy over classes. The mean squared error (MSE) is computed with respect to the average mIoU value and minimized by the introduction of Fairness; the same observation hold for standard deviation ($\sigma$) and entropy, which is however maximized.

\subsection{Performance} \label{sec:performance}
\rev{Equipping models with LEAK improves the segmentation performances of the model with only a slight additional occupancy in terms of RAM. For example, Cylinder3D \cite{zhou2020cylinder3d} with SemanticKITTI \cite{behley2019semantickitti} shows an increase in memory occupancy of $+28$ MB. This variation is almost negligible considering the size of a general point cloud segmentation network (\eg, Cylinder3D architecture takes around $8$ GB).
Moreover, inference time is not affected by our method. Training time is slightly worse for LEAK-based methods, but the change is almost negligible. For example, Cylinder3D with SemanticKITTI takes about $1.68$ seconds per iteration with LEAK, while about $1.61$ seconds per iteration without LEAK.}

\begin{table}[t]
\centering
\footnotesize
\caption{Analyses on SemanticKITTI \cite{behley2019semantickitti} and RandLA-Net \cite{hu2020randla}. Feature-prototype alignment measured by $\Theta_{\Gamma}$, CCD, PD and PCD.}
\label{tab:res:metrics} 
\setlength{\tabcolsep}{3.5pt}
\renewcommand{\arraystretch}{1.1}
\centering 
\begin{tabular}{r|c|c|c|c|c}
\toprule
   & \textbf{mIoU}           & $\Theta_{\Gamma}$ $\uparrow$ 
   & CCD $\downarrow$ & PD $\downarrow$ & PCD $\downarrow$  \\
   \midrule
Baseline        & 53.4            &  0.78 &	8.04 & 8.61 & 3.84 \\
\textbf{LEAK} & \textbf{54.7}  &  \textbf{0.89} &	\textbf{1.22} & \textbf{1.15} & \textbf{0.26}\\ 
\bottomrule
\end{tabular}
\end{table}

\begin{table}[t]
\centering
\footnotesize
\caption{Analyses on SemanticKITTI \cite{behley2019semantickitti} and RandLA-Net \cite{hu2020randla}. Class-balancing effect measured by standard deviation ($\sigma$), MSE and entropy.}
\label{tab:res:metrics2} 
\setlength{\tabcolsep}{3.5pt}
\renewcommand{\arraystretch}{1.1}
\centering 
\begin{tabular}{r|c|c|c|c}
\toprule
   & \textbf{mIoU}           & $\sigma$ $\downarrow$ & MSE $\downarrow$ & entropy $\uparrow$ \\ \midrule
Baseline        & 53.4            &  29.0 &	796.8	&  1.94 \\
Fairness & \textbf{54.4}  &  \textbf{28.2} &	\textbf{754.8}	& \textbf{1.97} \\
\bottomrule
\end{tabular} \vskip -1.5ex
\end{table}

\section{Conclusion}\label{sec:conclusions}
In this paper, we showed that accuracy, convergence time, and homogeneity can be improved for standard point cloud semantic segmentation methods, by automatically analyzing the classification mistakes of the original models. Thanks to these errors, we identify \textit{macro} classes, hierarchically grouping sets of mutually misclassified \textit{micro} classes.
Experimental evidence showed \textit{a posteriori} that such groups are consistent in terms of semantic characterization or classification difficulty. (\eg, due to the sample frequency or sparsity).
This information is exploited in model learning to regularize feature space in a two-fold manner. First, we cluster features of the same class (both macro and micro) tightly around their prototypical representations. Second, we constrained the class-wise accuracy score to be equally distributed across \textit{micro} classes contained within the same \textit{macro} cluster.
The proposed method boosted network performance while reducing the convergence time. Also, our solution proved to be totally agnostic to the backbone architecture and dataset, and remarkably prompt to generalization, as it was adapted to 4 different datasets, 3 architectures, and 2 types of input data (\ie, point clouds and RGB images).

\bibliographystyle{splncs04}
\bibliography{main}

\end{document}